%% file: ms.tex
\documentclass[a4paper]{article}

\usepackage{INTERSPEECH2019}
\usepackage{tabularx}
\usepackage{threeparttable}

\title{Personalizing ASR for Dysarthric and Accented Speech with Limited Data}
\name{Joel Shor$^1$, Dotan Emanuel$^1$, Oran Lang$^1$, Omry Tuval$^1$, Michael Brenner$^1$, Julie Cattiau$^1$, \\ Fernando Vieira$^2$, Maeve McNally$^2$, Taylor Charbonneau$^2$, Melissa Nollstadt$^2$, Avinatan Hassidim$^1$, Yossi Matias$^1$}
\address{
  $^1$Google, United States\\
  $^2$ALS Therapy Development Institute, United States}
\email{\{joelshor, dotan, oranl\}@google.com}

\begin{document}

\maketitle

\begin{abstract}
\input{sections/abstract.tex}
\end{abstract}
\noindent\textbf{Index Terms}: speech recognition, personalization, accessibility

\section{Introduction}
\input{sections/introduction.tex}

\section{Related Work}

\input{sections/related_works.tex}



\section{Experiments}

\input{sections/experiments.tex}

\section{Results}

\input{sections/results.tex}

\section{Discussion}

\input{sections/discussion.tex}

\section{Future Work}

\input{sections/future_work.tex}

\section{Acknowledgements}

We'd like to thank Tara Sainath, Anshuman Tripathi, Hasim Sak, Ding Zhao, Ron Weiss, Chung-Cheng Chiu, Dan Liebling,
and Philip Nelson for technical and project guidance.
\pagebreak

\bibliography{mybib} {}
\bibliographystyle{IEEEtran}

\end{document}

%% file: sections/abstract.tex
Automatic speech recognition (ASR) systems have dramatically improved over the last few years.
ASR systems are most often trained from ‘typical’ speech, which means that underrepresented groups don’t experience the same level of improvement.
In this paper, we present and evaluate finetuning techniques to improve ASR for users with non-standard speech.
We focus on two types of non-standard speech: speech from people with amyotrophic lateral sclerosis (ALS) and accented speech.
We train personalized models that achieve $62\%$ and $35\%$ relative WER improvement on these two groups,
bringing the absolute WER for ALS speakers, on a test set of message bank phrases, down to $10\%$ for mild dysarthria
and $20\%$ for more serious dysarthria.
We show that $71\%$ of the improvement comes from only 5 minutes of training data.
Finetuning a particular subset of layers (with many fewer parameters) often gives better results than finetuning the entire model.
This is the first step towards building state of the art ASR models for dysarthric speech.

%% file: sections/introduction.tex
State-of-the-art speaker-independent ASR systems are made possible by
large datasets and data-driven algorithms. This setup works well when the
dataset contains a large amount of data from all types of voices that one would
want the system to recognize, but fails on groups not well-represented in the data:
ASR models trained on
thousands of hours of `typical users' recognize voices in the typical use-case,
but often fail to recognize accented voices or ones impacted by medical conditions.

Getting ASR to work on non-standard speech is difficult for a few reasons. First, it can
be difficult to find enough speakers to train a state-of-the-art model. Second,
individuals within a group like `ALS' or a particular accent can have very different
ways of speaking. This paper's approach overcomes data scarcity by beginning with
a base model trained on thousands of hours of standard speech. It
gets around sub-group heterogeneity by training personalized models.

This paper details the technical approach to Project Euphonia, an accessibility
project announced at Google I/O  2019 \cite{euphoniavideo2019, euphoniablog2019}.
To demonstrate our approach, we focus on
accented speech from non-native English speakers,
and speech from individuals living with ALS. ALS is a progressive
neurodegenerative disorder that affects speech production, among other things.
About $25\%$ of people with ALS experience slurred speech as their first symptom \cite{ALSTDI}.
Most people with ALS eventually lose mobility, so they would especially
benefit from being able to interact verbally with smart home devices.
The severity of dysarthria for people with ALS is measured with an FRS score,
ranging from 0 to 4, with 0 being incomprehensible and 4 normal \cite{frs_def}.
In this paper, we improve state-of-the-art ASR for individuals with ALS FRS 1-3 and heavy
accents. The main contributions of this paper are:

\begin{enumerate}
\item A finetuning technique for personalizing an ASR model on dysarthric speech that yields
word error rates on a test set of message bank phrases \cite{BostonChildrensHospital}
of $10.8\%$ for mild dysarthria (FRS of 3) and $20.9\%$ for severe dysarthria (FRS of 1-2).
The finetuning technique yields a word error rate of $8.5\%$ on heavily accented data.
The finetuned models make recognition mistakes that are distributionally more
similar to standard ASR mistakes on standard speech.

\item Demonstrating that the finetuning gives significant improvement over the base model
in multiple scenarios, including different non-standard speech (dysarthric and accented), and
on different architectures.

\item On the ALS speech, 71\% of the relative WER improvement can be achieved
with 5 minutes of data, 80\% with 10 minutes of data. Furthermore, for some models,
just training the encoder produces better results, and
just training the layers closest to the input yields $90\%$ of total relative improvement.
\end{enumerate}

%% file: sections/related_works.tex
Neural networks are the state-of-the-art systems for ASR in the large data-regime. We explore
how well two particular architectures can be finetuned on a small amount of non-standard
data. The RNN-Transducer \cite{RNNT2012, RNNT2013} is a neural network
architecture that has shown good results on numerous ASR tasks. It consists
of an encoder and decoder network and is configured
such that unidirectional models can perform streaming ASR. In this paper, we use
a bidirectional encoder without attention that was shown to achieve comparable results \cite{Seq2SeqComparison2017}.

We also explore the Listen, Attend, and Spell architecture. It is an attention-based, sequence-to-sequence
model that maps sequences of acoustic properties to sequences of language
\cite{Chan2015}. It has produced state-of-the-art results on a challenging
12,500 hour voice-search ASR task, achieving a $4.1\%$ WER \cite{CC2018}. This
model uses an encoder to convert the sequence of acoustic frames to a sequence
of internal representations, and a decoder with attention to convert the sequence
of internal representations to linguistic output. The best network in \cite{CC2018}
produced word pieces, which are a linguistic representation between graphemes
and words \cite{WPM}.

There are a number of methods for adapting large ASR models to
small amounts of data \cite{DomainAdaptation2018,
MultiBasisAdaptive2015, DomainAdaptation2011}. This paper's approach is most similar to \cite{Bansal2018},
which involves selectively finetuning parts of the ASR model.

There are also many published techniques on how to improve ASR specifically for
pathological, or dysarthric, speech:
\cite{ArticulatoryASR} appended articulatory features to the usual acoustic ones
to achieve a 4-8\% relative WER improvement, and \cite{DysarthricASR2014} adapts ASR
models trained on open source datasets to a dataset of dysarthric speech. However,
all these approaches are limited either by the quality of the base model or the
amount of data available for finetuning.

Numerous studies have explored the acoustic, articulatory, and phonetic differences
between standard and modified speech. Some of the conditions explored are
Parkinsons \cite{Parkisons2018}, age \cite{OldAge2018}, dyslexia \cite{Dyslexia2018},
and ALS \cite{Glottal2018}. We add to this body of work by describing the phonetic
mistakes that a production ASR system makes on a large collection of ALS audio.
We are also able to describe our improved ASR model by characterizing which
phonemes the improved models are better able to recognize.

%% file: sections/experiments.tex
\begin{figure}[t]
  \centering
  \includegraphics[width=0.4\textwidth]{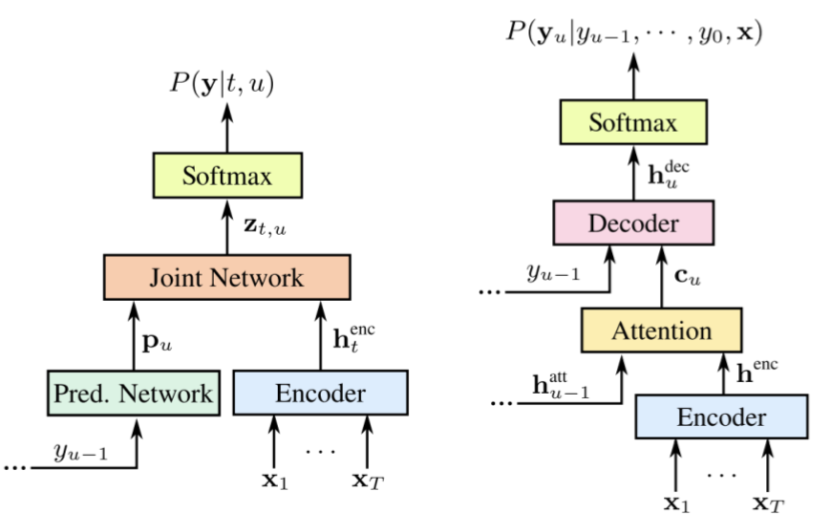}
  \caption{Schematic diagrams of the RNN-T architecture (left) and the LAS
    architecture (right). Diagrams come from \cite{Seq2SeqComparison2017}.}
  \label{fig:model_architecture}
\end{figure}

We create personalized ASR models by starting with a base model trained
on standard, unaccented speech. This approach is much more resource efficient
than retraining the entire model, from scratch, for each speaker.
In order to verify that we are learning something other than
just the idiosyncrasies of a particular model, we run most of our experiments starting from
two different models: a Bidirectional RNN Transducer (RNN-T) model \cite{RNNT2012},
and a Listen, Attend, and Spell (LAS) model \cite{Chan2015}.
Both are end-to-end sequence-to-sequence models. We follow the training procedure
in \cite{Seq2SeqComparison2017} for the RNN-T model and \cite{CC2018} for the LAS model
(on the 1000 hour, open source Librispeech dataset \cite{Librispeech}).

We finetune different layer combinations on different amounts of data, on both ALS and accents datasets. We
finetune models per speaker.

\subsection{Data}

\subsubsection{ALS}

  We collected $36.7$ hours of audio from $67$ people with ALS, in partnership with the
  ALS Therapy Development Institute (ALS-TDI). The participants were given sentences to read,
  and they recorded themselves on their home computers using custom software. The
  sentences were collected from three sources: The Cornell Movie-Dialogs Corpus
  \cite{CornellMovieCorpus}, a collection of sentences used by text-to-speech
  voice actors, and a modified selection of sentences from the Boston Children's Hospital \cite{BostonChildrensHospital}.
  Note that this corpus is a restricted language domain,
  but is phonetically very similar to other corpora e.g. Librispeech.
  The FRS scores of participants were measured by ALS-TDI, and we only evaluate
  on people with speech FRS 3 and below ($17$ speakers, $22.1$ hours). See
  the attached multimedia file for audio examples.

\subsubsection{Accented Speech}

  To test our finetuning method on another type of non-standard speech, we use the
  L2 Arctic dataset of non-native speech \cite{L2Arctic}. This dataset consists of
  $20$ speakers with approximately 1 hour of speech per speaker. Each speaker recorded
  a set of 1150 phonetically balanced utterances. For each of the 20 speakers, we split the data
  into 90/10 train and test. All of the sentences which contains proper nouns
  are used in the training set, in order to remove the possibility of the model to
  artificially achieve better results on the test set by memorizing them.

\subsection{Base Models}

All our base networks and finetuning are trained on 80-bin log-mel spectrograms computed
from a 25ms window and a 10ms hop. We use the same technique presented in \cite{Hasim2016}
and used in \cite{Seq2SeqComparison2017}: we stack frames in groups of 3 and process
them as one 'super-frame.'

We trained RNN-T and LAS architectures (Figure \ref{fig:model_architecture}).
All training was performed with the multicondition training (MTR) techniques described
in \cite{Seq2SeqComparison2017}. During training, we distorted the audio using
a room simulator derived from YouTube data. The average SNR of the added noise is 12dB.
We use the TensorFlow library Lingvo \cite{shen2019lingvo}.

\subsubsection{Bidirectional RNN-Transducer}

In this paper, we primarily work with a bidirectional RNN-Transducer (RNN-T)
architecture that achieves near state-of-the-art performance. This architecture
was first introduced in \cite{RNNT2012}. We use the version presented in \cite{Seq2SeqComparison2017}.
The network maps acoustic frames to word pieces, which are a linguistic representation between graphemes
and words \cite{WPM}. It has a 5 layer bidirectional convolutional LSTM encoder, a 2 layer LSTM decoder, and a joint layer.
It has 49.6M parameters in total.

\subsubsection{Listen, Attend, and Spell model}

To verify which results generalize beyond a particular architecture,
we also run some of our experiments with a Listen, Attend, and Spell (LAS) model
trained on the open source Librispeech dataset \cite{Librispeech}.
This architecture was first introduced in \cite{Chan2015}.
The model that we use is described in \cite{CC2018}. It is a sequence-to-sequence with
attention model that maps acoustic frames to graphemes. We use an encoder with 4 layers of bidrectional convolutional LSTMs
and a 2 layer RNN decoder. The model has a total of 132M parameters.

Our grapheme targets are the 26 English lower-case letters, punctuation symbols, and a space.
There are 33 target grapheme symbols. The base LAS model was trained to 1M steps
on all 960 hours of the Librispeech dataset.
It achieved a $5.5\%$ WER on the clean test split and $15.5\%$ on the
non-clean test split (the two standard test splits given in the dataset).

\subsection{Finetuning}

All our finetuning uses four Tesla V100 GPUs for no more than four hours.

\subsubsection{RNN-T}

We started by finetuning 1, 2, and 3 layers in fixed combinations
(treating the decoder as a single layer), on both
datasets, adjusting hyperparameters as necessary.
Let E$_i$ denote the $i$th layer of the encoder, where lower-numbered layers are
closer to the input.
We uniformly found that training from E$_0$ up with or without the
joint layer was always better than the other methods, so we focused our search
on training from E$_0$ up, with or without the joint layer.

Next, for each ALS and accented individual, we exhaustively searched our finetuning
space, with various amounts of data. For the RNN-T, this meant finetuning
each of E$_0$, E$_0$-E$_1$, E$_0$-E$_2$, etc.. the entire encoder, with or without the joint layer.

\subsubsection{LAS}

For LAS architectures, we finetuned various layer combinations and consistently found that
the best results from this network came from finetuning all layers. All
results reported in this paper for the LAS network are on finetuning
the entire network, unless otherwise specified.

%% file: sections/results.tex
\begin{table}[t]
  \centering
  \begin{threeparttable}[b]
  \caption{Average WER Improvements}
  \label{table:abs_wer_nums}
  \begin{tabularx}{\columnwidth}{X X X X X X}
    \toprule
    \multicolumn{1}{l}{} & \multicolumn{1}{l}{\textbf{Cloud}} & \multicolumn{2}{c}{\textbf{RNN-T}} & \multicolumn{2}{c}{\textbf{LAS}} \\
    \midrule
    \multicolumn{2}{l}{}                                        & \multicolumn{1}{c}{Base} & \multicolumn{1}{c}{Finetune}  & \multicolumn{1}{c}{Base} & \multicolumn{1}{c}{Finetune}  \\
    \midrule
    Arctic\tnote{1} &   \centering $24.0$  &  \centering $13.3$    &  \centering $8.5$  &  \centering $22.6$    & $11.3$               \\
    ALS\tnote{2}    &   \centering $42.7$  &  \centering $59.7$    &  \centering $20.9$ &  \centering $86.3$    & $31.3$                \\
    ALS\tnote{3}    &   \centering $13.1$  &  \centering $33.1$    &  \centering $10.8$ &  \centering $49.6$    & $17.2$                \\
    \bottomrule
  \end{tabularx}
  \begin{tablenotes}
       \item[1] Non-native English speech from the L2-Arctic dataset. \cite{L2Arctic}
       \item[2] Low FRS (ALS Functional Rating Scale) intelligible with repeating, Speech combined with nonvocal communication.
       \item[3] FRS-3 detectable speech disturbance. \cite{frs_def}
      \end{tablenotes}
  \end{threeparttable}
\end{table}

\begin{figure}[t]
  \centering
  \includegraphics[width=0.4\textwidth]{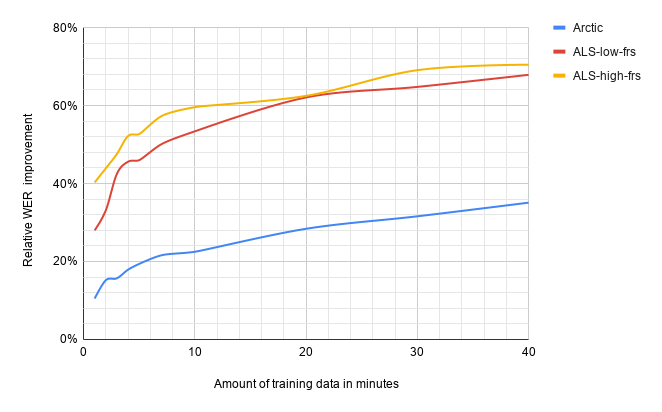}
  \caption{Average Relative WER improvement as a function of the amount of
  training data.}
  \label{fig:rel_improvement}
\end{figure}

\subsection{Performance}

We report our absolute word error rate in Table \ref{table:abs_wer_nums}. The results
show dramatic improvement over Google cloud ASR \cite{CloudASR} model for very non-standard speech
(heavy accents and ALS speech below 3 on the ALS Functional Rating Scale \cite{frs_def})
and moderate improvements in ALS speech that is similar to healthy speech. The comparison
to Google cloud ASR demonstrates that a healthy speech model finetuned on
non-standard speech produces strong results, but we acknowledge that
the comparison is not perfect: the LAS and RNN-T models use the entire
audio to make predictions while Google cloud ASR model only looks backwards in time. This lets Google cloud ASR
support streaming, but it is also less accurate.

The relative difference between base model and
the finetuned model demonstrates that the majority of the improvement comes from the finetuning
process, except in the case of the RNN-T on the Arctic dataset (where the RNN-T
baseline is already strong).

\subsection{Limited Data}

On the ALS dataset, finetuning on five minutes and ten minutes of data yields
$75\%$ and $85\%$ of the WER improvement compared to a model trained on ~40 min of data, respectively.
On the Arctic dataset, fourteen and twenty minutes of training data yields
$70\%$ and $81\%$ of the WER improvement, respectively.
Figure \ref{fig:rel_improvement}  shows more details.

\subsection{Layers}

The LAS model consistently performed best when the entire network was finetuned.
The RNN-T model achieved $91\%$ of the relative WER improvement by just finetuning
the joint layer and first layer of the encoder (compared to finetuning the joint layer and the
entire encoder, the average WER across all participants
regardless of FRS score was 18.1\% vs 15.1\%). On Arctic, finetuning the joint layer and the first encoder
layer achieved $86\%$ of the relative improvement compared to finetuning the
entire network ($11.0\%$ vs $10.5\%$).

\subsection{Phoneme mistakes}

\begin{figure}[t]
  \centering
  \includegraphics[width=0.4\textwidth]{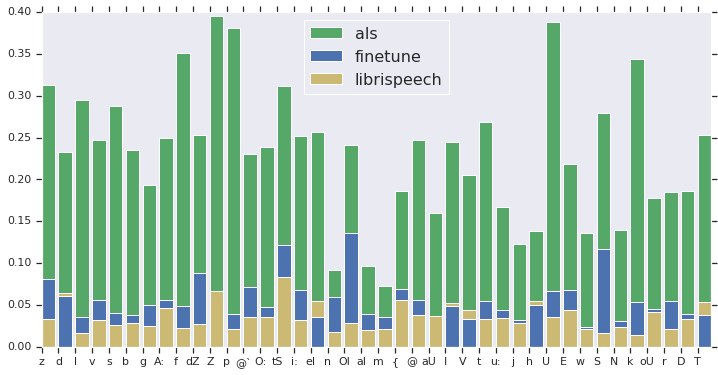}
  \caption{The distribution of phoneme mistakes before and after finetuning. x-axis
    is SAMPA phoneme. y-axis is number of times that phoneme was deleted or substituted
    divided by number of phoneme occurances in the ground truth transcripts.}
  \label{fig:dist_improvement}
\end{figure}

To better understand how our models improved, we looked at the pattern
of phoneme mistakes. We started by comparing the distribution of phoneme mistakes
made by Google cloud ASR model on standard speech (Librispeech) to the mistakes made on ALS speech.
We map the ground truth transcripts and the ASR model outputs to sequences of SAMPA phonemes,
then compute the edit distance between the two phoneme sequences. We aggregate
over speakers and utterances.

First, we compute the probability of
a mistake for a particular phoneme in the ground truth transcript. Since the
standard speech and ALS speech transcripts are different, we normalize by the
number of ground truth transcript phonemes. The phonemes
with the five largest differences between
the ALS data and standard speech are p, U, f, k, and Z. These five account for
$20\%$ of the likelihood of a deletion mistake. Next, we
investigate which phonemes are mistakenly added. We compute the
probability that a particular phoneme will mistakenly appear in the
recognized transcript. By far the biggest differences are in the n and m phonemes,
which together account for 17\% of the insertion / substitution mistakes.

Finally, we perform the same analysis on some of our finetuned models.
The unrecognized phoneme distribution becomes more similar to that
of standard speech (see Figure \ref{fig:dist_improvement}). Also, surprisingly,
the phoneme distribution that the model produced when it made a mistake was much more
similar to mistakes on standard speech after finetuning
(KL of $0.26$ between standard and ALS, $0.10$ after finetuning).

%% file: sections/discussion.tex
In this paper, we develop well-performing ASR models for dysarthric and
heavily accented individuals by carefully finetuning healthy-speech models.
Specifically, we demonstrated:
\begin{enumerate}
\item Good absolute performance on average dysarthric ALS speakers, large improvements in very dysarthric speakers
\item Better performance on ALS and accented speech when just training the RNN-T encoder
\item Much of the improvement is from the first 5-10 minutes of training data, and
can be achieved by just training the first encoder layer and the joint layer
\item The five most mistaken phonemes in ALS speech account for 20\% of the
mistakes
\end{enumerate}

Prior to this work, it was unclear how much improvement could be achieved by
finetuning on small amounts of non-standard speech. We show that with on the order of 1 hour of data, we can create
a personalized ASR model that is significantly better than Cloud-based services. 
This improvement comes
mainly from the finetuning process: we see a relative WER improvement over
our base model of $70\%$ for dysarthric speech (for both groups) and
$35.1\%$ for accented speech. The finetuning
process used four GPUs for fewer than four hours, making this technique very
accessible.

We achieve better performance by just training the encoder for
the RNN-T architecture but not the LAS architecture. This is likely due to a peculiarity of the RNN-T:
the RNN-T architecture is factorized into a component whose activations
are solely a function of the current audio (the encoder) and another component whose activations
are a function of the current predicted transcript (the decoder).
The LAS model has no such factorization, so information about the acoustics and linguistics
are likely more evenly distributed throughout the network. This claim isn't
precise, since the RNN-T weights used at inference time are determined by the statistical
properties of both language and acoustics, but even minor effects from this factorization
might mean that finetuning the encoder only helps prevent the network
from overfitting to the language seen in the small amount of training data.

A large fraction of the improvement comes from the first 5-10 minutes of audio.
One explanation is that the test and train sentences are linguistically similar enough
for the model to learn the kinds of things that are said during testing from
a small number of examples. Another explanation is that just finetuning part of the network
allows it to retain the general acoustic and linguistic information from the general speech
model while needing minimal modifications to adapt to a single new speaker. Future
work includes testing this hypothesis, possibly by exploring its performance on
radically different sentences.

The model might be adapting to general non-standard speech or to the individual.
This could be tested, for instance, by
training a single model on the entire ALS or Arctic corpus, and comparing it
to a single speaker model. We didn't include results from
this comparison, and we discuss this in the `Future Works' section.

We found that finetuning just the joint layer and the first encoder layer
achieved $90\%$ of the relative improvement compared to training the joint layer
and the entire encoder. This can be explained by a combination of the acoustic
vs linguistic properties discussed earlier in this section and by analogy with
a popular computer vision finetuning technique. Many computer vision papers
publish good results by finetuning the last few layers of a classification
network that was pretrained on Imagenet \cite{ImagePretraining}. In that technique, authors
usually assume that their data follows roughly the distribution of images in imagenet,
and that their image label distribution is different. Our problem has the reverse assumption:
we assume that the distribution of language (labels) is the same in our problem as in standard
speech, but that the distribution of ALS / accented speech is different from that of standard
speech.

Lastly, we find that the five most deleted / substituted phonemes account for
roughly 20\% of such errors, and that the two most incorrectly inserted phonemes
account for just over 20\% of such errors. This kind of observation might lead to
ALS-detection techniques: one can try to detect ALS degradation by matching
the prediction distribution of a production ASR system. We might also be able to improve
the finetuning process by collecting more speech that involves the most-often-confused
phonemes.

%% file: sections/future_work.tex
A major challenge is to build state of the art speech recognition models for strongly dysarthric speech.
It is an open question whether there are additional techniques that can be helpful
in the low data regime (such as Virtual Adversarial Training, data augmentation, etc).
We can also use the phoneme mistakes to
weight certain examples during training, or to pick training sentences for people with
ALS to record that contain the most egregious phoneme mistakes.

We would like to explore pooling data from multiple speakers with similar conditions,
but did not do so in this paper. We believe that such an experiment
raises more questions around training speaker-independent non-standard speech
ALS models, which we feel are outside the scope of this work.